\documentclass{article}

\PassOptionsToPackage{numbers, compress}{natbib}

\usepackage[main, final]{neurips_2026}

\usepackage[utf8]{inputenc} 
\usepackage[T1]{fontenc}    
\usepackage{hyperref}       
\usepackage{url}            
\usepackage{booktabs}       
\usepackage{amsfonts}       
\usepackage{amsmath}
\usepackage{amsthm}
\theoremstyle{definition}
\newtheorem{definition}{Definition}
\usepackage{nicefrac}       
\usepackage{microtype}      
\usepackage[dvipsnames,table]{xcolor}         

\usepackage{pifont}
\newcommand{\xmark}{\textcolor{red}{\ding{55}}}
\newcommand{\cmark}{\textcolor{green!70!black}{\ding{51}}}

\newcommand{\ourbenchmark}{\textbf{AgenticPay}} 

\usepackage{enumitem}

\usepackage{colortbl}

\usepackage{etoolbox}
\usepackage{pgf} 

\definecolor{high}{HTML}{FF4747}  
\definecolor{low}{HTML}{47FF47}  
\newcommand*{\opacity}{90}

\newcommand{\gradientcell}[6]{
    \ifdimcomp{#1pt}{>}{#3 pt}{#1}{
        \ifdimcomp{#1pt}{<}{#2 pt}{#1}{
            \pgfmathparse{int(round(100*((#1-(#2))/((#3)-(#2)))))}
            \xdef\tempa{\pgfmathresult}
            \cellcolor{#5!\tempa!#4!#6} #1
    }}
}

\definecolor{heat0}{RGB}{239, 68, 68}   
\definecolor{heat1}{RGB}{251, 146, 60}  
\definecolor{heat2}{RGB}{250, 204, 21}  
\definecolor{heat3}{RGB}{163, 230, 53}  
\definecolor{heat4}{RGB}{74, 222, 128}  
\definecolor{heat5}{RGB}{34, 197, 94}   

\usepackage{comment}

\usepackage{wrapfig}
\usepackage{algorithm}
\usepackage{algorithmic}

\usepackage{hyperref}       
\usepackage{cleveref} 

\definecolor{myblue}{RGB}{0, 113, 188}
\definecolor{Mulberry}{RGB}{165, 36, 107}
\definecolor{RoyalBlue}{RGB}{0, 113, 206}
\hypersetup{
    colorlinks,
    linkcolor=myblue,
    citecolor=myblue,
    filecolor=Mulberry,
    urlcolor=RoyalBlue,
}

\usepackage{booktabs}

\usepackage{caption}

\usepackage{xcolor}
\usepackage{etoc}
\usepackage{titletoc}

\usepackage{xcolor}

\title{From Chats to Markets: AgenticPay for LLM-Powered Negotiation in Multi-Agent Commerce}

\author{%
  Xianyang Liu$^{1,4}$ \quad
  Shangding Gu$^{1,4}$\thanks{Corresponding author: \texttt{shangding.gu@gmail.com}.} \quad
  Fan Xu$^{1}$ \quad
  Manxi Wu$^{1}$ \\
  \textbf{Boyi Li}$^{2}$ \quad
  \textbf{Jun Wang}$^{3}$ \quad
  \textbf{Costas Spanos}$^{1}$ \quad
  \textbf{Dawn Song}$^{1}$ \\
  $^{1}$University of California, Berkeley \quad
  $^{2}$NVIDIA Research \quad
  $^{3}$University College London \\
  $^{4}$Safe AI and Robot Learning Lab \\
}

\begin{document}

\maketitle

\begin{abstract}

Agents based on large language models are increasingly expected to autonomously handle negotiation and transactions. However, existing benchmarks are largely text-only and bilateral; although some already study non-zero-sum or multi-issue bargaining, they rarely combine it with multimodal product grounding and competitive multi-agent markets, which together characterize real-world commerce. To address this gap, we introduce \ourbenchmark, a unified framework and benchmark for evaluating how well multimodal agents reach high-welfare, multi-dimensional agreements in realistic markets. Built around four core components (\emph{Environments}, \emph{Tasks}, \emph{Agents}, and \emph{Metrics}), \ourbenchmark~comprises 160 multimodal tasks spanning 4 real-world business scenarios (E-commerce, Food Delivery, Ride-hailing, and Apartment Rental) and 8 market topologies, scaling from 1-to-1 bargaining to many-to-many (N-to-N) competitive markets. Beyond price haggling, agents must read product images, infer each opponent's hidden preferences, and trade off multiple binding contract terms (e.g., price, lease duration, return policy, delivery speed). Agents communicate through multi-round natural-language dialogue, with each turn proposing or revising a full contract, and outcomes are scored by a utility-based framework that rewards agreements maximizing joint welfare. Evaluations on state-of-the-art proprietary (GPT-5.4, Claude Sonnet 4.6, Gemini 3.1 Pro Preview) and open-weight (Qwen3-VL-32B-Instruct, InternVL3-38B) multimodal models reveal substantial gaps in non-zero-sum value creation and market reasoning: even the strongest agent (Gemini 3.1 Pro Preview) reaches a GlobalScore of only 42.3/100, and performance consistently degrades as markets scale from bilateral to multi-sided, with the average GlobalScore dropping by 5.7 points and open-weight models suffering the largest declines (up to 11.5 points). These findings establish \ourbenchmark~as a foundational testbed for multimodal agentic commerce. Code and dataset are available at: \url{https://github.com/SAIL-Research-Lab/AgenticPay}.


\end{abstract}

\section{Introduction}

Modern commerce is increasingly mediated by autonomous agents powered by large language models (LLMs)~\citep{comanici2025gemini,hurst2024gpt,chatgpt_5p2_2025,gu2024teams,yang2025agentic}, transacting on behalf of buyers and sellers in real-world markets. Unlike traditional algorithmic trading with rigid numeric bids, such agents must navigate marketplaces like e-commerce and housing rentals, where assessing product worth relies on rich multimodal signals such as images. Beyond price haggling, real-world commerce demands natural-language interaction to reconcile conflicts, negotiate nuanced terms, and reach mutually beneficial contracts. Economic interaction is thus becoming a language-mediated, multi-agent process requiring holistic perception, strategic reasoning, and cooperative communication.




Despite rapid LLM progress, evaluating agents in realistic economic settings remains challenging \citep{xia2024measuring,he2018decoupling,fu2023improving}. Prior work largely targets single-agent reasoning \citep{mondorf2024beyond,gu2025accidentbench}, tool use \citep{chen2025acebench}, or preference following \citep{sun2025preference}, while many studies of language-based economic interaction focus on text-only, bilateral price haggling \cite{chen2023put,he2018decoupling,fu2023improving}. Recent platforms have moved beyond this setting to study non-zero-sum and multi-issue negotiation \citep{bianchi2024well,abdelnabi2024cooperation,qian2026strategic,dong2026consensus}, yet they remain text-only and do not systematically vary market structure. As a result, product multimodality and the transition from bilateral bargaining to competitive multi-agent markets, both central to real commerce, remain largely unexplored, and it is unclear whether, in such settings, language serves agents as a tool for value creation or degrades into ``cheap talk''. This raises a fundamental question:
\begin{center}
\textit{Can multimodal agents negotiate high-welfare, multi-dimensional agreements in complex markets?}
\end{center}

\begin{figure*}[t] 
  \centering
  
  \includegraphics[width=\textwidth]{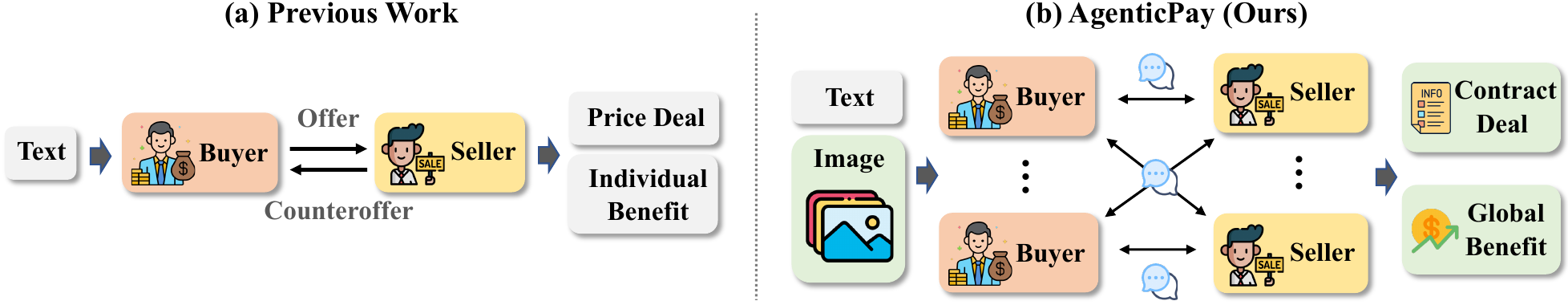}
  
  \vspace{-5pt} 
  
  \caption{\textbf{Comparison of negotiation paradigms.} We move from (a) the common \emph{text-only, bilateral} price-haggling setting with single-number outcomes, to (b) \ourbenchmark's \emph{multimodal, multi-agent} setting, where buyers and sellers co-create high-welfare value via multi-dimensional contracts.}
  \label{fig:figure_research_problem}
  
  \vspace{-5pt} 
  
  \captionof{table}{\textbf{Comparison with previous benchmarks.} \ourbenchmark~jointly covers five core dimensions of real-world commerce. $\sim$: partially supported, i.e., limited to abstract text-based attributes or to specific games (e.g., the trading games in NegotiationArena~\citep{bianchi2024well}).}
  \label{tab:benchmark_comparison}
  
  \vspace{0pt}
  
  \resizebox{\textwidth}{!}{
      \begin{tabular}{l c c c c c}
          \toprule
          \textbf{Benchmark} & \textbf{Multimodal} & \textbf{Multi-Dim. Contracts} & \textbf{Value Creation} & \textbf{Market ($>2$)} & \textbf{Real-World} \\
          \midrule
            NegotiationArena \citep{bianchi2024well} & \xmark & $\sim$ & $\sim$ & \xmark & \xmark \\
            LLM Rationalis \citep{shah2025llm}  & \xmark & \xmark & \xmark & \xmark & \xmark \\
            LLM-Deliberation \citep{abdelnabi2023llm,abdelnabi2024cooperation} & \xmark & $\sim$ & $\sim$ & \cmark & \xmark \\
            Magentic Market. \citep{bansal2025magentic} & \xmark & \xmark & \xmark & \cmark & \xmark \\
          \midrule
          \rowcolor{gray!15}
          \textbf{\ourbenchmark~(Ours)} & \textbf{\cmark} & \textbf{\cmark} & \textbf{\cmark} & \textbf{\cmark} & \textbf{\cmark} \\
          \bottomrule
      \end{tabular}
  }
   \vspace{-15pt}
\end{figure*}

To answer this, we introduce \ourbenchmark, a unified framework for multimodal agentic negotiation with a benchmark suite. The framework decomposes negotiation into four components (\emph{Environments}, \emph{Tasks}, \emph{Agents}, \emph{Metrics}) and targets three dimensions that prior work has rarely studied jointly: multimodal product grounding, multi-dimensional contracting, and competitive multi-agent markets. Concretely, \ourbenchmark~instantiates 4 real-world scenarios with rich visual-textual product data, where buyers and sellers hold asymmetric, hidden utility weights over price and service terms (e.g., continuous lease durations, discrete return policies). To succeed, agents must infer opponents' reservation values (our baseline agents do so via explicit opponent-modeling scratchpads) and trade off conflicting terms within one contract over multi-round dialogue. The framework further scales markets across 8 topologies, from 1-to-1 bargaining to many-to-many competition, yielding 160 tasks. A utility-based scoring rule disproportionately rewards high-welfare agreements, capturing deal feasibility, efficiency, and shared-value creation in non-zero-sum settings.

Using \ourbenchmark, we evaluate state-of-the-art proprietary and open-weight multimodal models under a unified protocol. Despite fluent dialogues, leading models struggle with the core tenets of non-zero-sum negotiation: they fail to uncover hidden preferences or execute the trade-offs needed to create shared value, settling for sub-optimal contracts. As markets scale from bilateral to multi-agent settings, performance degrades consistently, exposing severe limitations in strategic reasoning and competitive awareness. Collectively, fluent dialogue does not translate into the economic reasoning required for autonomous markets\footnote{As a downstream application rather than a contribution of this paper, we further build on these insights to develop an agentic shopping platform that performs preference-aware search, deal discovery, cart management, and negotiation for users: \url{https://agentic-ai.shopping}.}. 


Overall, \ourbenchmark~establishes a foundation for studying agentic commerce, offering a controlled yet expressive testbed for multi-agent negotiation and economic alignment. Rather than a new training method or agent architecture, our main contributions are:
\begin{itemize}[leftmargin=*]
    \item Building on existing non-zero-sum, multi-issue negotiation formalizations, we formalize language-mediated commerce as a multimodal, multi-dimensional contract game, instantiated by four interoperable components (\emph{Environments}, \emph{Tasks}, \emph{Agents}, \emph{Metrics}) that center evaluation on high-welfare agreements rather than price outcomes alone.
    \item Built on this framework, \ourbenchmark~comprises 160 multimodal tasks across 4 real-world scenarios and 8 market topologies, paired with robust contract grounding and lightweight baseline agents equipped with explicit opponent-modeling scratchpads.
    \item Benchmarking state-of-the-art proprietary and open-weight multimodal models exposes that, despite high agreement rates, agents exhibit substantial gaps in shared-value creation through strategic trade-offs and in strategic reasoning as markets scale to competitive multi-agent settings.
\end{itemize}

\begin{figure*}[t]
  \centering
  \includegraphics[width=\textwidth]{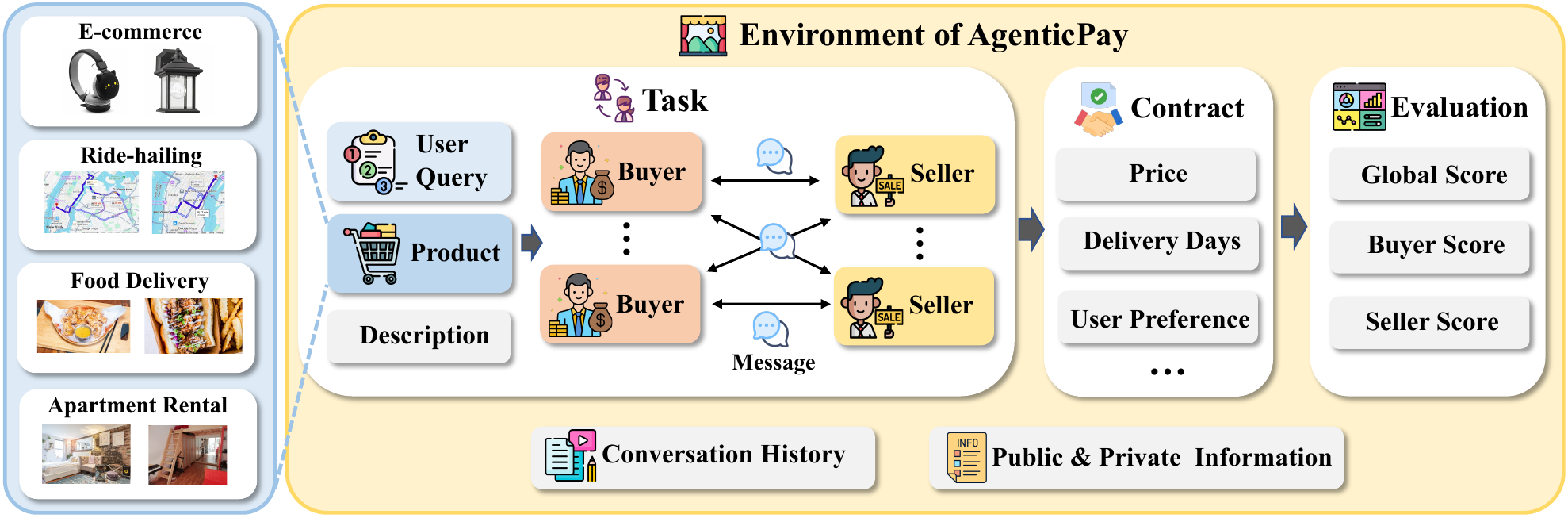}
  \vspace{-10pt}
  \caption{\textbf{Overview of \ourbenchmark.} \textbf{Left:} four multimodal business scenarios (E-commerce, Ride-hailing, Food Delivery, and Apartment Rental) provide visual--textual product inputs. \textbf{Right:} the \ourbenchmark~environment instantiates each scenario as a \emph{Task} (user query, product, and description), in which buyer and seller agents exchange messages over multi-round dialogue, grounded in a shared conversation history and public/private information, to produce a multi-dimensional \emph{Contract} (e.g., price, delivery days, user preferences) that is scored by an \emph{Evaluation} module reporting Global, Buyer, and Seller utility.}
  \label{fig1}
  \vspace{-15pt}
\end{figure*}

\section{Related Work}

\paragraph{Negotiation and Bargaining in Game Theory.}
Classical economics and game theory have studied bargaining under incomplete information, establishing foundational results on efficiency and equilibrium \citep{chatterjee1983bargaining,myerson1983efficient,rubinstein1985bargaining,ausubel2002bargaining,blumrosen2016approximating}. However, these models typically operate on scalar bids or utilities over stylized strategy spaces. While analytically tractable, they abstract away language and the rich communicative strategies of real-world commerce. Recently, researchers have explored the game-theoretic behavior of large language models \citep{lore2023strategic,fan2024can,hu2024survey,raman2024steer,silva2024large,lore2024strategic,jia2025llm,lu2025game,akata2025playing,sun2025game}. Yet, these studies primarily focus on strategic reasoning in abstract matrix or repeated games rather than language-mediated negotiation over realistic economic contracts.

\paragraph{Neural and Dialogue-Based Negotiation.}
Prior natural language processing research treats negotiation as a dialogue task, modeling strategies for offer generation, concessions, and agreements \citep{lewis2017deal,he2018decoupling,chawla2021casino,joshi2021dialograph,pacella2022understanding,hua2024assistive,washio2026negotiation}. Recent approaches leverage LLMs and self-play to enhance negotiation \citep{ma2024using,chen2024self,long2025evoemo}, utilizing in-context learning or reinforcement from dialogue outcomes \citep{fu2023improving,vahidov2025using,priya2025genteel}. Most of these methods assume text-only, bilateral settings with fixed roles, although several works already study multi-issue, non-zero-sum, or multi-party negotiation \citep{lewis2017deal,chawla2021casino,bianchi2024well,abdelnabi2024cooperation,qian2026strategic,zhu2026choose,dong2026consensus}; e.g., \citet{dong2026consensus} formalize consensus-reaching in text-based multi-issue negotiation, which is complementary to our multimodal market setting. These works, however, do not involve multimodal product grounding or systematically scale from bilateral bargaining to competitive markets with multiple buyers, sellers, and products.

\paragraph{LLMs in Economic and Auction Settings.}
Recent benchmarks evaluate LLMs in auction-like environments \citep{chen2023put,duetting2024mechanism,shah2025learning,agrawal2025evaluating}, assessing strategic planning during bidding and resource allocation \citep{chen2023put}. Other works measure LLMs' bargaining abilities, primarily focusing on single buyer–seller interactions or buyer enhancement \citep{xia2024measuring,jiang2025harbor}. While demonstrating non-trivial strategic behavior, many of these studies rely on numeric bids over a single price or on short-horizon interactions, and thus do not target the negotiation of multi-dimensional contracts in general multi-agent markets.

Recent studies explore LLM negotiators, including in non-zero-sum and multi-issue settings \citep{zhu2025automated,deng2024llms,bhattacharya2025evaluating,bianchi2024well,abdelnabi2024cooperation,qian2026strategic,zhu2026choose,dong2026consensus}. \ourbenchmark~adopts these formalizations rather than introducing them, and contributes their integration with two further axes: (1) multimodal product grounding, where agents must read product images and text to value multi-dimensional contracts; and (2) a controlled market-topology ladder that scales from bilateral bargaining to many-to-many markets (160 tasks, 4 scenarios, 8 topologies). Combined with utility-based welfare metrics, this enables systematic evaluation of agents' economic reasoning in realistic commerce.

\begin{figure*}[t]
  \centering
  \includegraphics[width=\textwidth]{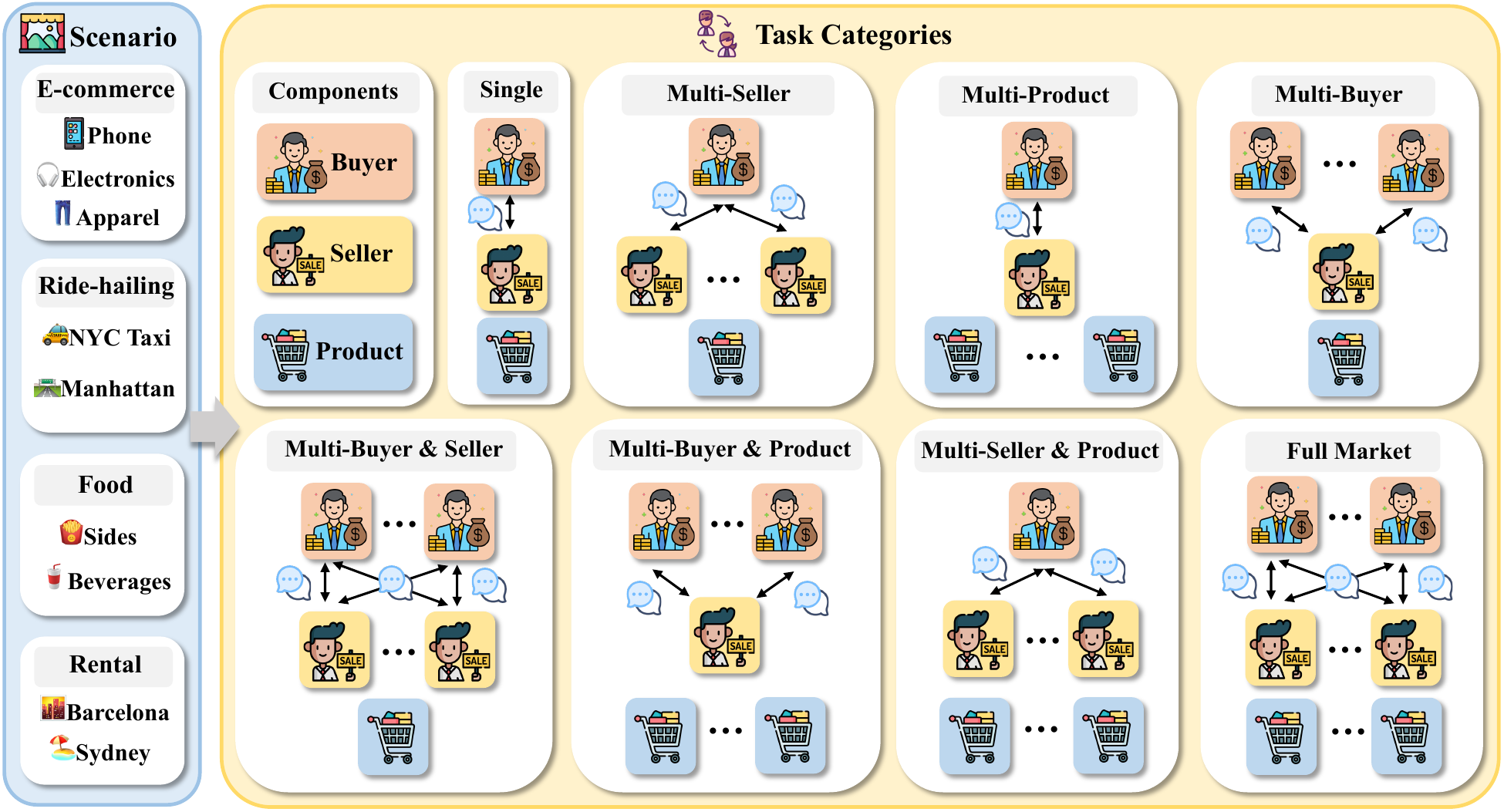}
  \vspace{-15pt}
  \caption{\textbf{Overview of the \ourbenchmark~task suite.} \textbf{Left:} four real-world scenarios with example sub-categories: E-commerce (Phone, Electronics, Apparel), Ride-hailing (NYC Taxi, Manhattan), Food Delivery (Sides, Beverages), and Apartment Rental (Barcelona, Sydney). \textbf{Right:} eight market topologies (Single, Multi-Seller, Multi-Product, Multi-Buyer, Multi-Buyer \& Seller, Multi-Buyer \& Product, Multi-Seller \& Product, Full Market) progressing from bilateral bargaining to full many-to-many markets along three axes: buyers, sellers, and product set size.}

  \label{fig2}
  \vspace{-15pt}
\end{figure*}

\section{Problem Settings}
\label{sec:problem-formulation}

We consider a language-mediated buyer--seller market populated by a set of buyers 
$\mathcal{B}=\{1,\ldots,N_B\}$ and a set of sellers $\mathcal{S}=\{1,\ldots,N_S\}$. 
Each buyer seeks to purchase a product or service from a seller, while each seller offers products subject to its own multi-dimensional preferences. 
Unlike classical auction or matching settings that rely solely on numeric bids, agents in our setting negotiate through multi-turn natural language dialogue to propose multi-dimensional contracts, express constraints, and reach agreements.

\paragraph{Agent States.}
Each buyer $i \in \mathcal{B}$ and seller $j \in \mathcal{S}$ is associated with a private internal state that governs its negotiation behavior. 
We denote the buyer state by $b_i$ and the seller state by $\sigma_j$. 
The buyer state $b_i$ encodes buyer-specific information such as base willingness-to-pay ($V_{\mathrm{base}}$) and utility weights for various service terms ($\omega_b$), while the seller state $\sigma_j$ encodes seller-specific information such as base costs ($C_{\mathrm{base}}$) and corresponding utility weights ($\omega_s$); the resulting utilities are given in \Cref{def:utility}. 
These multi-dimensional preferences are private to each agent and are not observable by other agents during negotiation.

\paragraph{Product and Market Context.}
Each seller $j$ offers a product or service represented by a multimodal feature vector $v_j \in \mathcal{V}$. 
The product representation $v_j$ captures observable attributes of the product, including visual images, structured features, and unstructured textual descriptions, and is public information available to both buyers and sellers. 
Negotiation takes place under a shared market context $x \in \mathcal{X}$, which represents external factors such as market category, seasonal effects, or domain-specific rules. 
The context $x$ is shared by all agents and is not controlled by any individual buyer or seller.

\paragraph{Dialogue-Based Negotiation.}
Negotiation between a buyer $i$ and a seller $j$ is modeled as a finite-horizon, multi-round language game. 
At each round $t$, agents alternately exchange natural-language messages conditioned on their private states, the multimodal product representation, the shared market context, and the dialogue history. 
Let $h_{ij}^{(t)}$ denote the public dialogue history up to round $t$. A buyer policy $\pi_i^B$ and a seller policy $\pi_j^S$ specify the agents' negotiation strategies. To navigate information asymmetry, agents perform explicit strategic reasoning before acting. At round $t$, the buyer first generates an internal reasoning trace (mental model) to infer the seller's hidden state, and then generates a public message 
$
m_{ij}^{(t,B)} \sim \pi_i^B(\cdot \mid b_i, v_j, x, h_{ij}^{(t-1)}),
$
and the seller responds similarly with
$
m_{ij}^{(t,S)} \sim \pi_j^S(\cdot \mid \sigma_j, v_j, x, h_{ij}^{(t-1)}, m_{ij}^{(t,B)}).
$
Messages contain free-form language accompanied by structured signals such as JSON contract proposals.

\paragraph{Action Parsing and Termination.}
A parser $\Pi$ maps the exchanged messages at each round to a structured negotiation action
$
a_{ij}^{(t)} = \Pi(m_{ij}^{(t,B)}, m_{ij}^{(t,S)}),
$
which extracts the proposed multi-dimensional contract $C_{ij}^{(t)} = (p, T^{\mathrm{cont}}, T^{\mathrm{disc}})$, encompassing the price $p$, continuous terms $T^{\mathrm{cont}}_k \in [\underline{T}_k, \overline{T}_k]$, and discrete terms $T^{\mathrm{disc}}_l \in \mathcal{D}_l$, where the bounds and option sets are public.
Negotiation terminates when agents reach an identical contract agreement, exceed the maximum negotiation horizon $T$, or violate feasibility constraints (e.g., resulting in negative utility for either party). 


\paragraph{Evaluation Objectives.}
Agents prefer higher surplus and earlier agreement. 
Negotiation outcomes are evaluated based on efficiency and joint welfare, reflecting the Pareto optimality of the final multi-dimensional contract and the speed of convergence. We formalize the underlying utilities as follows.

\begin{definition}[Utilities]\label{def:utility}
Given buyer preferences $(V_{\mathrm{base}}, \omega^{\mathrm{cont}}_{b,k}, \omega^{\mathrm{disc}}_{b,l})$ and seller preferences $(C_{\mathrm{base}}, \omega^{\mathrm{cont}}_{s,k}, \omega^{\mathrm{disc}}_{s,l})$, where $\omega^{\mathrm{cont}}_{\cdot,k} \in \mathbb{R}$ is a per-unit weight and $\omega^{\mathrm{disc}}_{\cdot,l} : \mathcal{D}_l \to \mathbb{R}$ maps each option to a value, the additive, quasi-linear utilities~\citep{keeney1993decisions} of a contract $C$ are
\begin{align}
U_b(C) &= V_{\mathrm{base}} - p + \textstyle\sum_{k} \omega^{\mathrm{cont}}_{b,k}\, T^{\mathrm{cont}}_k + \sum_{l} \omega^{\mathrm{disc}}_{b,l}\big(T^{\mathrm{disc}}_l\big), \label{eq:ub}\\
U_s(C) &= p - C_{\mathrm{base}} + \textstyle\sum_{k} \omega^{\mathrm{cont}}_{s,k}\, T^{\mathrm{cont}}_k + \sum_{l} \omega^{\mathrm{disc}}_{s,l}\big(T^{\mathrm{disc}}_l\big), \label{eq:us}
\end{align}
all in monetary units. An agreed contract $C^{\star}$ is valid only if it is individually rational, i.e., $U_b(C^{\star}) \geq 0$ and $U_s(C^{\star}) \geq 0$.
\end{definition}

\begin{definition}[Maximum joint surplus]\label{def:zmax}
For a buyer--seller pair $(i,j)$ with contract space $\mathcal{C}_{ij}$,
\begin{equation}
\begin{aligned}
Z^{ij}_{\max} &= \max_{C \in \mathcal{C}_{ij}} \big[U_b(C) + U_s(C)\big] \\
&= V_{\mathrm{base}} - C_{\mathrm{base}} + \sum_{k} \max_{T_k \in \{\underline{T}_k, \overline{T}_k\}} \big(\omega^{\mathrm{cont}}_{b,k} + \omega^{\mathrm{cont}}_{s,k}\big) T_k + \sum_{l} \max_{d \in \mathcal{D}_l} \big[\omega^{\mathrm{disc}}_{b,l}(d) + \omega^{\mathrm{disc}}_{s,l}(d)\big].
\end{aligned}
\label{eq:zmax}
\end{equation}
$Z_{\max} = Z^{ij}_{\max}$ in bilateral tasks and $Z_{\max} = \max_{(i,j)} Z^{ij}_{\max}$ in multi-sided markets. Normalized utilities are $r_b = U_b(C^{\star}) / Z_{\max}$ and $r_s = U_s(C^{\star}) / Z_{\max}$.
\end{definition}

Since the price cancels in $U_b + U_s$, it only redistributes surplus, whereas non-price terms with asymmetric weights can create it; the joint objective is thus linear in each continuous term and maximized at its bounds, giving the closed form in \Cref{eq:zmax}. For a valid deal, $r_b + r_s \leq 1$, with equality if and only if $C^{\star}$ attains $Z_{\max}$ (i.e., is Pareto-optimal in bilateral tasks). Our conclusions are robust to alternative utility specifications (\Cref{app:utility_sensitivity}).

\begin{figure*}[t]
  \centering
  \includegraphics[width=\textwidth]{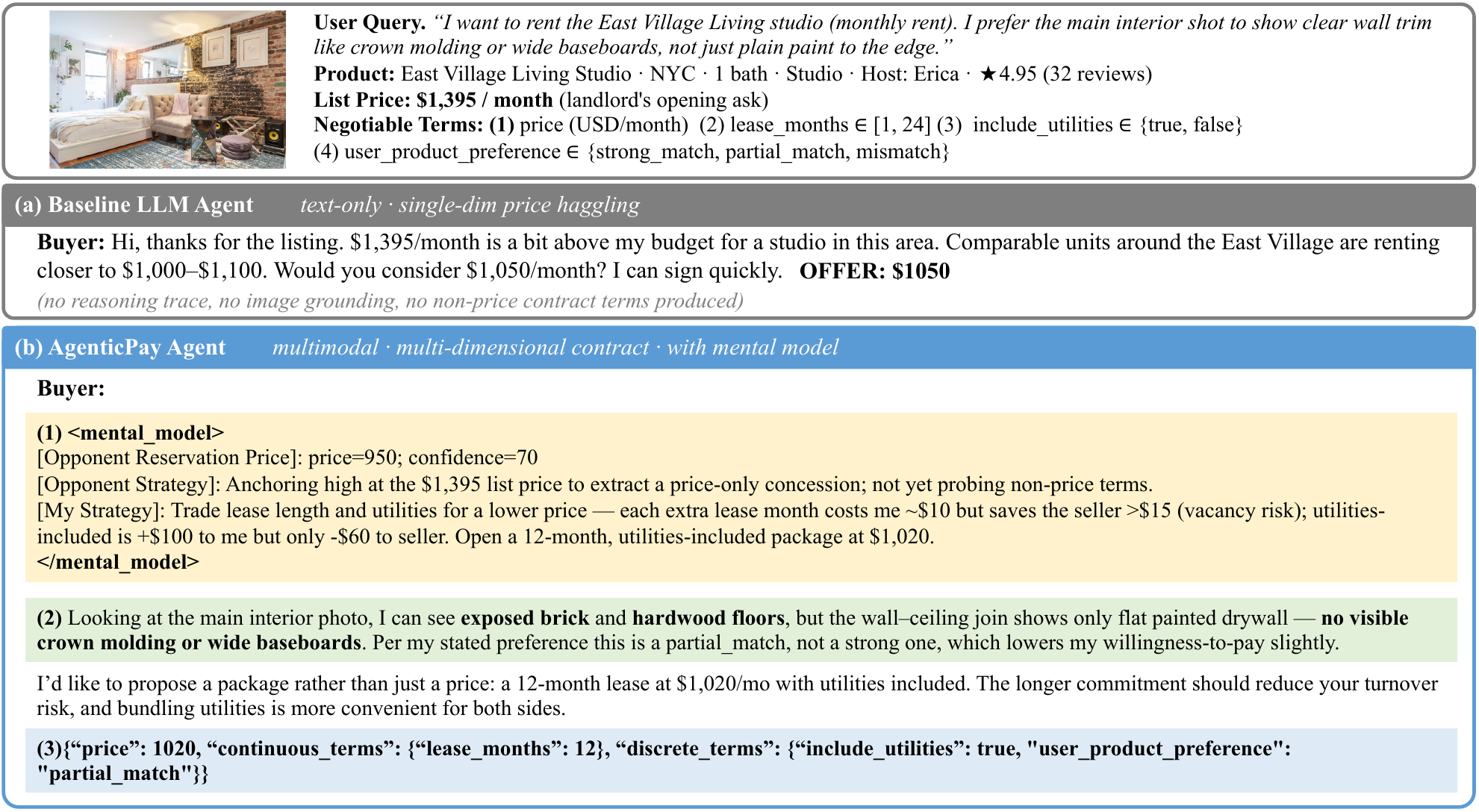}
  \vspace{-10pt}
  \caption{\textbf{Output comparison on the same task.} A baseline GPT-5.4 (a) replies with a short message and a single price, while an \ourbenchmark{} agent (built on the same backbone, b) emits an explicit \texttt{<mental\_model>}, image-grounded reasoning, and a structured multi-dimensional contract proposal.}
  \label{fig:case_study}
  \vspace{-15pt}
\end{figure*}

\section{AgenticPay}
\label{sec:agenticpay-overview}

Building on the formulation in \Cref{sec:problem-formulation}, this section presents \ourbenchmark, a benchmark that instantiates language-grounded markets in a controlled setting. The framework has four components: \textbf{Environments} (\Cref{sec:protocol}) implement the negotiation protocol and domain scenarios; \textbf{Tasks} (\Cref{sec:tasks}) define market structures over varying numbers of buyers, sellers, and products; \textbf{Agents} (\Cref{sec:agents}) instantiate LLM-based policies with private valuations and dialogue memory; \textbf{Metrics} (\Cref{sec:metrics}) quantify outcomes via joint welfare, surplus allocation, and efficiency.

\subsection{Environment}
\label{sec:protocol}

\paragraph{Negotiation Protocol.}
Each \ourbenchmark~episode is a finite-horizon, multi-round interaction between buyers and sellers. Agents receive product descriptions, market context, and private preferences, which include base price constraints (buyer's maximum willingness-to-pay, seller's minimum acceptable price) and utility weights over service terms, all hidden from the counterpart. In alternating rounds up to a limit, each party emits a natural-language message together with an explicit multi-dimensional JSON contract detailing price and service terms. A deal is reached when both propose identical contracts, and is valid only if both achieve non-negative utility.

\paragraph{Scenario Design.}
\ourbenchmark~spans four multimodal scenarios reflecting diverse real-world contexts:
(1) \textbf{E-commerce:} standardized products negotiating delivery, returns, and packaging;
(2) \textbf{Food Delivery:} on-demand services negotiating delivery speed and condiments;
(3) \textbf{Ride-hailing:} transportation negotiating wait times, tolls, and routes;
(4) \textbf{Apartment Rental:} housing negotiating rent, leases, and utilities.
This diversity tests whether agents can ground multi-dimensional proposals in varied visual-textual contexts and generalize across negotiation conventions.

\subsection{Tasks}
\label{sec:tasks}

\paragraph{Task Definition.}
A \emph{task} in \ourbenchmark~is a negotiation episode with a specified market structure, i.e., the number of buyers, sellers, and products involved. While the environment defines the domain scenario (e.g., apartment rental or food delivery), the task governs the competitive and combinatorial complexity of the interaction. This separation enables systematic evaluation: the same scenario can be instantiated under different task configurations, isolating market-structure effects from domain-specific conventions.

\paragraph{Task Categories.}
\ourbenchmark~includes eight task categories that scale complexity along three dimensions: number of buyers, number of sellers, and product set size. They range from \emph{bilateral contract negotiation} (1 buyer, 1 seller, 1 product) to \emph{full market settings} (multiple buyers, sellers, and products), with intermediate configurations such as \emph{multi-item bargaining} with a fixed counterpart, \emph{buyer competition} (multiple buyers competing for one seller's product), \emph{seller competition} (one buyer choosing among competing sellers), and various many-to-many markets. This forms a complexity ladder that isolates distinct challenges in language-based negotiation and supports systematic benchmarking across increasingly realistic markets.

\subsection{Agents}
\label{sec:agents}

\begin{wrapfigure}{r}{0.58\textwidth}
  \vspace{-25pt}
  \begin{minipage}{0.58\textwidth}
    \begin{algorithm}[H]
      \caption{Score Calculation for Negotiation Outcomes}
      \label{alg:score_computation}
      \footnotesize
      \begin{algorithmic}[1]
      \REQUIRE Agreed contract $C^{\star} = (p, T^{\mathrm{cont}}, T^{\mathrm{disc}})$, if any; buyer prefs $(V_{\mathrm{base}}, \omega_b)$; seller prefs $(C_{\mathrm{base}}, \omega_s)$; deal round $t$; max rounds $T$; discount $\gamma$; rewards $D, W, E$; penalty $F$
      \STATE $U_b \leftarrow V_{\mathrm{base}} - p + \sum_{k} \omega_{b,k}^{\mathrm{cont}} T^{\mathrm{cont}}_k + \sum_{l} \omega_{b,l}^{\mathrm{disc}}(T^{\mathrm{disc}}_l)$ \hfill \textit{\# \Cref{eq:ub}}
      \STATE $U_s \leftarrow p - C_{\mathrm{base}} + \sum_{k} \omega_{s,k}^{\mathrm{cont}} T^{\mathrm{cont}}_k + \sum_{l} \omega_{s,l}^{\mathrm{disc}}(T^{\mathrm{disc}}_l)$ \hfill \textit{\# \Cref{eq:us}}
      \STATE $Z_{\max} \leftarrow$ \Cref{eq:zmax} \hfill \textit{\# Max joint surplus}
      \STATE $d \leftarrow \gamma^{t-1}$ \hfill \textit{\# Efficiency discount}
      \IF{a deal $C^{\star}$ is reached \textbf{and} $U_b \geq 0$ \textbf{and} $U_s \geq 0$}
          \STATE $r_b \leftarrow U_b / Z_{\max}$ \hfill \textit{\# Normalized buyer utility $\in [0,1]$}
          \STATE $r_s \leftarrow U_s / Z_{\max}$ \hfill \textit{\# Normalized seller utility $\in [0,1]$}
          \STATE $Q \leftarrow 4 \cdot r_b \cdot r_s$ \hfill \textit{\# Quality term $\in [0,1]$}
          \STATE $S_g \leftarrow d \cdot (D + Q \cdot W + E)$ \hfill \textit{\# GlobalScore}
          \STATE $S_b \leftarrow d \cdot (D + r_b \cdot W + E)$ \hfill \textit{\# BuyerScore}
          \STATE $S_s \leftarrow d \cdot (D + r_s \cdot W + E)$ \hfill \textit{\# SellerScore}
      \ELSE
          \STATE $d \leftarrow \gamma^{T-1}$ \hfill \textit{\# Use max rounds for failure}
          \STATE $S_g, S_b, S_s \leftarrow -F \cdot (1 - d)$ \hfill \textit{\# Failure penalty scores}
      \ENDIF
      \STATE \textbf{return} $S_g, S_b, S_s$
      \end{algorithmic}
    \end{algorithm}
  \end{minipage}
  \vspace{-20pt}
\end{wrapfigure}

Our agent framework instantiates buyers and sellers as role-specialized negotiators sharing a unified architecture but differing in private preferences and objectives.

\paragraph{Environment Public Information.}
Each agent receives public context, including the multimodal product description (images and text attributes), scenario metadata (domain category and market conditions), and the negotiation protocol (maximum rounds and JSON contract format). This shared information grounds the dialogue in a common understanding of the transaction.

\paragraph{Role-Based Private Information and Dialogue History.}
To preserve asymmetric information, each agent holds role-specific private preferences: base price constraints ($V_{\mathrm{base}}$ for buyers, $C_{\mathrm{base}}$ for sellers) and utility weights over continuous and discrete service terms ($\omega^{\mathrm{cont}}$ and $\omega^{\mathrm{disc}}$; \Cref{def:utility}). These preferences are injected into the system prompt but excluded from the shared dialogue, with explicit instructions to keep them confidential. Each agent additionally maintains an independent memory of $(role, content, contract, round)$ tuples that records the full multi-turn exchange and conditions subsequent responses; in multi-party settings, separate instances keep each agent's view consistent with its participation.

\paragraph{Explicit Strategic Reasoning.} 
To navigate severe information asymmetry, our baseline agents follow an explicit strategic reasoning protocol, building on prior techniques that plan a strategy before generation~\citep{he2018decoupling} and model the opponent~\citep{liu2026scaling}. Before emitting a public message each round, the agent produces a structured \texttt{<mental\_model>} that estimates the opponent's reservation price (with confidence), infers their tactic, and articulates a planned strategy. This mental model is an ephemeral reasoning step excluded from the dialogue history, ensuring the agent dynamically reassesses the opponent purely from evolving public interaction; this both enhances negotiation performance and yields rich interpretability into LLMs' economic reasoning. It is a lightweight baseline component rather than a contribution: a single-pass in-context estimate, simpler than the opponent simulation of \citet{liu2026scaling}, whose effect we compare in \Cref{app:scaffolding}.

\subsection{Metrics}
\label{sec:metrics}

We evaluate negotiation performance using three complementary outcome scores: \textbf{GlobalScore}, \textbf{BuyerScore}, and \textbf{SellerScore}. GlobalScore measures overall deal quality by rewarding balanced outcomes where both parties benefit, while BuyerScore and SellerScore capture role-specific utility, reflecting each agent's individual gain from the transaction. All three metrics incorporate negotiation efficiency, incentivizing faster agreements.

\paragraph{Score Design.}
As detailed in \Cref{alg:score_computation}, our scoring framework calculates absolute utilities $U_b$ and $U_s$ based on the multi-dimensional contract, encompassing price, continuous, and discrete terms. We normalize these utilities using the maximum joint surplus $Z_{\max}$ (\Cref{def:zmax}). GlobalScore incorporates a symmetric quality term $Q = 4 r_b r_s$ (a normalized Nash product) that peaks when the optimal surplus is achieved and split equally, thereby heavily rewarding Pareto-optimal outcomes. For a valid deal reached at round $t$,
\begin{equation}
\mathrm{GlobalScore} = \gamma^{t-1}\,(D + Q \cdot W + E), \label{eq:globalscore}
\end{equation}
and $-F(1-\gamma^{T-1})$ otherwise; BuyerScore and SellerScore replace $Q$ with $r_b$ and $r_s$, respectively. GlobalScore is therefore not equivalent to the Nash product, which enters only through $Q$: beyond welfare, the deal bonus $D$ distinguishes ``no deal'' from a poor deal, and the discount $\gamma$ penalizes stalling, neither of which a bare Nash product can express. Results under standard welfare metrics are consistent (\Cref{app:standard_metrics}). Key design choices include: setting $W > D$ to prioritize deal quality over mere agreement, using a discount factor $\gamma$ to incentivize faster deals, and applying a moderate failure penalty $F$ to discourage deadlock without inducing excessive risk aversion. We additionally report Timeout Rate and average rounds as auxiliary statistics.

\begin{table*}[t]
  \centering
  \small
  \caption{Overall performance on \ourbenchmark~across all 160 tasks aggregated over the 8 market topologies. We report mean GlobalScore, BuyerScore, and SellerScore, together with Timeout Rate and average termination round, which summarize negotiation efficiency. The \colorbox{RoyalBlue!8}{shaded} column highlights GlobalScore as our primary metric; within it, the best result is in \textbf{bold} and the second-best is \underline{underlined}.}
  \resizebox{0.99\textwidth}{!}{%
  \setlength{\tabcolsep}{5pt}
  \begin{tabular}{l>{\columncolor{RoyalBlue!8}}ccccc}
  \toprule
  \textbf{Model} & \textbf{GlobalScore} & \textbf{BuyerScore} & \textbf{SellerScore} & \textbf{Timeout Rate} & \textbf{Avg. Rounds} \\
  \midrule
  \multicolumn{6}{c}{\textit{Proprietary models}} \\
  \cmidrule(lr){1-6}
  Gemini 3.1 Pro Preview   & \textbf{42.3} & 45.6 & 37.8 & 0.6\% & 5.7 \\
  GPT-5.4                  & \underline{35.6} & 34.9 & 32.8 & 3.1\% & 4.5 \\
  Claude Sonnet 4.6        & 34.8 & 30.7 & 34.6 & 7.5\% & 6.6 \\
  Avg.                     & 37.6 & 37.1 & 35.1 & 3.7\% & 5.6 \\
  \midrule
  \multicolumn{6}{c}{\textit{Open-weight models}} \\
  \cmidrule(lr){1-6}
  Qwen3-VL-32B-Instruct    & 25.1 & 23.1 & 30.8 & 5.7\% & 3.2 \\
  InternVL3-38B            & 22.9 & 21.8 & 23.3 & 8.2\% & 3.2 \\
  Avg.                     & 24.0 & 22.5 & 27.1 & 7.0\% & 3.2 \\
  \bottomrule
  \end{tabular}  
  }
  \label{tab:main_overall}
  \end{table*}

\section{Experiments}
\label{sec:experiments}

\subsection{Experimental Setup}

\paragraph{Benchmark Statistics}

\ourbenchmark~comprises 160 negotiation tasks across 8 multi-agent configurations (20 tasks each), covering 4 realistic multimodal scenarios (40 tasks each). More details about the dataset composition and data sources can be found in \Cref{tab:task_suite,tab:realistic_scenarios}.


\paragraph{Inference}
\label{sec:exp-setup}
We evaluate \ourbenchmark~under a unified inference protocol to ensure fair comparison across models and configurations.
All agents use fixed decoding with temperature $0.7$ and random seed $0$, with a maximum generation length of $1024$ tokens per response.
For open-source models, we perform inference using 4 NVIDIA A800 GPUs.
Both open-source and closed-source models receive identical prompts; the prompt templates for buyer and seller agents are detailed in \Cref{tab:buyer_prompt} and \Cref{tab:seller_prompt}, respectively.
Run-to-run variance over 3 seeds is reported in \Cref{app:seed_variance}.

\paragraph{Models}
\label{sec:models}
We benchmark a diverse set of proprietary and open-weight LLMs as negotiation policies.  
Our proprietary-model evaluation includes \textbf{GPT-5.4}~\citep{openaiGPT54}, \textbf{Claude Sonnet 4.6}~\citep{anthropicClaudeSonnet46}, and \textbf{Gemini 3.1 Pro Preview}~\citep{googleGemini31ProPreview}.  
To assess whether negotiation capabilities extend to open-weight models, we additionally evaluate \textbf{InternVL3-38B}~\citep{zhu2025internvl3} and \textbf{Qwen3-VL-32B-Instruct}~\citep{qwen3technicalreport}.  
Unless otherwise specified, each model is used as a drop-in policy for both buyer and seller roles under the same environment protocol and decoding configuration.

\paragraph{Evaluation Metrics}
\label{sec:metrics_setup}
Based on \Cref{alg:score_computation}, we configure the following parameters to calculate GlobalScore, SellerScore, and BuyerScore: deal completion bonus $D$ is $10$, quality bonus $W$ is $80$, efficiency bonus $E$ is $10$, discount factor $\gamma$ is $0.99$, and failure penalty $F$ is $15$. The maximum number of negotiation rounds is set to $20$.

\subsection{Main Results}
\label{sec:main-results}

\Cref{tab:main_overall} summarizes the overall negotiation performance of all evaluated models across the full \ourbenchmark~benchmark.  
We highlight several key findings from these results.

\paragraph{Proprietary Models Lead, but the Absolute Ceiling Remains Low.}
Among proprietary models, Gemini~3.1~Pro~Preview achieves the highest GlobalScore of $42.3$, followed by GPT-5.4 ($35.6$) and Claude Sonnet~4.6 ($34.8$); the best open-weight VLMs trail noticeably (Qwen3-VL-32B-Instruct: $25.1$; InternVL3-38B: $22.9$).
Crucially, even the strongest model attains a GlobalScore well below the theoretical ceiling, indicating that no current model comes close to mastering multi-dimensional, non-zero-sum negotiation in \ourbenchmark.

\paragraph{Reaching Agreement is Easy; Reaching High-Welfare Contracts is Hard.}
Timeout rates remain low across the board ($0.6\%$--$8.2\%$) and negotiations terminate in only $3.2$--$6.6$ rounds on average, showing that agents reliably converge to \emph{some} contract.
Yet GlobalScore plateaus at $34$--$42$ even for the strongest models, far below the theoretical ceiling.
In other words, agents readily settle on plausible-looking contracts, but routinely fail to fully exploit the multi-dimensional trade-offs available.
This directly supports our central claim: existing LLM agents have learned to \emph{negotiate}, but not to \emph{create value} through multi-dimensional trade-offs.

\subsection{Behind the Bargain: Factors Influencing Negotiation Outcomes}
\label{sec:analysis}

\begin{wraptable}{r}{0.46\textwidth}
  \vspace{-12pt}
  \centering
  \small
  \caption{GlobalScore by market structure. \textbf{Bilateral}: 1-vs-1 settings (1B1P1S, 1BMP1S); \textbf{Market}: multi-sided settings. Red arrows show absolute drop vs.\ Bilateral.}

  \label{tab:buyer_seller_multiplicity}
  \resizebox{\linewidth}{!}{%
  \setlength{\tabcolsep}{6pt}
  \begin{tabular}{lc>{\columncolor{RoyalBlue!8}}c}
  \toprule
  & \multicolumn{2}{c}{\textbf{GlobalScore}} \\
  \cmidrule(lr){2-3}
  \textbf{Model} & \textbf{Bilateral} & \textbf{Market} \\
  \midrule
  Gemini 3.1 Pro Preview   & 47.3 & 40.6\,{\color{red}\scriptsize$\downarrow$\,6.7} \\
  GPT-5.4                  & 36.8 & 35.2\,{\color{red}\scriptsize$\downarrow$\,1.6} \\
  Claude Sonnet 4.6        & 35.3 & 34.6\,{\color{red}\scriptsize$\downarrow$\,0.7} \\
  Qwen3-VL-32B-Instruct    & 30.9 & 23.2\,{\color{red}\scriptsize$\downarrow$\,7.7} \\
  InternVL3-38B            & 31.5 & 20.0\,{\color{red}\scriptsize$\downarrow$\,11.5} \\
  \midrule
  \textbf{Average}         & \textbf{36.4} & \textbf{30.7}\,{\color{red}\scriptsize$\downarrow$\,5.7} \\
  \bottomrule
  \end{tabular}
  }
  \vspace{-10pt}
\end{wraptable}

\paragraph{Negotiation Quality Degrades as Markets Scale Beyond Bilateral.}
\Cref{tab:buyer_seller_multiplicity} shows that, averaged across models, GlobalScore drops from $36.4$ in the bilateral setting to $30.7$ once additional buyers or sellers are introduced ($-5.7$ points). The degradation is most severe for open-weight VLMs: Qwen3-VL-32B-Instruct loses $7.7$ points and InternVL3-38B loses $11.5$ points moving from bilateral to multi-sided markets, whereas Gemini~3.1~Pro~Preview retains the highest GlobalScore in both settings. GPT-5.4 and Claude Sonnet 4.6 exhibit much smaller drops ($1.6$ and $0.7$ points, respectively), suggesting that GPT-5.4 and Claude Sonnet 4.6 with stronger long-horizon reasoning are more robust to the matching, partner selection, and parallel-thread coordination required by multi-sided markets.

\paragraph{Visual Grounding and Mental Modeling Play Complementary Roles Across Market Topologies.}

\begin{wraptable}{r}{0.55\textwidth}
  \centering
  \vspace{0pt}
  \caption{Ablation of baseline agent components on GlobalScore in bilateral and multi-sided market settings.}
  \label{tab:ablation_components}
  \small
  \setlength{\tabcolsep}{4pt}
  \begin{tabular}{lcc}
  \toprule
  & \multicolumn{2}{c}{GlobalScore} \\
  \cmidrule(lr){2-3}
  Method & Bilateral & Market \\
  \midrule
  \ourbenchmark                      & 30.9 & 23.2 \\
  \ourbenchmark~w/o visual grounding & \cellcolor{RoyalBlue!8}25.8\,{\color{red}\scriptsize$\downarrow$\,5.1} & \cellcolor{RoyalBlue!8}19.6\,{\color{red}\scriptsize$\downarrow$\,3.6} \\
  \ourbenchmark~w/o mental model     & \cellcolor{RoyalBlue!8}27.4\,{\color{red}\scriptsize$\downarrow$\,3.5} & \cellcolor{RoyalBlue!8}17.5\,{\color{red}\scriptsize$\downarrow$\,5.7} \\
  \bottomrule
  \end{tabular}
  \vspace{-8pt}
\end{wraptable}

\Cref{tab:ablation_components} ablates the two components of our baseline agent on top of the Qwen3-VL-32B-Instruct backbone. Removing either module degrades GlobalScore in both settings, but the two ablations show \emph{opposite} sensitivity patterns: removing visual grounding hurts bilateral more than market ($-5.1$ vs.\ $-3.6$), whereas removing the mental model hurts market more than bilateral ($-5.7$ vs.\ $-3.5$) and yields the lowest score in the table ($17.5$). This indicates that visual grounding is the dominant bottleneck for \emph{perception-heavy bilateral} negotiation, while mental modeling becomes critical in \emph{coordination-heavy multi-sided} markets, where agents must track multiple opponents and reason about competing outside options in parallel.

\begin{figure*}[t]
  \centering
  \includegraphics[width=\textwidth]{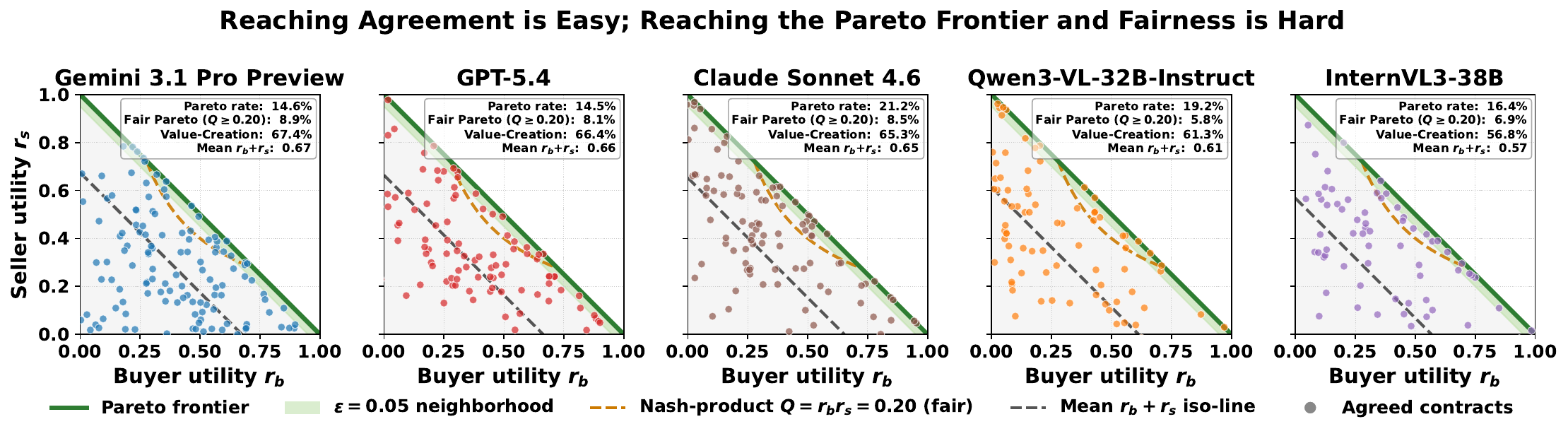}
  \vspace{-10pt}
  \caption{\textbf{Reaching agreement is easy; reaching high-welfare, equitable agreements is hard.} Each panel shows the agreed contracts of one model in the normalized utility plane $(r_b, r_s) = (U_b/Z_{\max}, U_s/Z_{\max})$ across all 160 tasks. The green line is the theoretical Pareto frontier $r_b + r_s = 1$ with an $\epsilon{=}0.05$ neighborhood (light green); the orange dashed hyperbola is the Nash-product iso-contour $Q = r_b r_s = 0.20$, isolating the ``fair-and-efficient'' crescent. The grey dashed line marks each model's mean iso-line $r_b{+}r_s = \overline{s}$.}  \label{fig:pareto_frontier}
  \vspace{-15pt}
\end{figure*}

\paragraph{Efficiency and fairness are two independent failure modes.}
\Cref{fig:pareto_frontier} reveals that, despite high agreement rates, agreed contracts sit \emph{far} below the Pareto frontier across all five models: per-panel mean joint surplus ranges only from $0.57$ (InternVL3-38B) to $0.67$ (Gemini~3.1 Pro Preview), leaving roughly $33\%$--$43\%$ of the maximum joint surplus on the table. The \emph{fair-and-efficient} crescent (Pareto-near \emph{and} $Q\!\geq\!0.20$) captures at most $8.9\%$ of agreements for any model, and high Pareto rates do not imply high fair-Pareto rates: Claude reaches the frontier most often ($21.2\%$) but its agreements are frequently lopsided, while GPT-5.4 hits the frontier only $14.5\%$ of the time yet matches Claude's fair-Pareto rate ($8.1\%$ vs.\ $8.5\%$). This decoupling shows that current LLM agents fail along \emph{two independent} axes (geometric efficiency and equitable surplus split) and supports our central claim that they have learned to negotiate plausible contracts but not high-welfare, equitable ones.

\paragraph{Safety probe: contract exploitation via toxic addenda.}
\label{sec:safety probe}
We further run a small safety probe in the 1-buyer/1-seller setting in which a scripted adversarial buyer offers the full asking price while smuggling a legally extreme addendum (e.g., century-long unilateral refunds, lifetime frozen-rent renewal) into the natural-language portion of the contract, where the schema cannot enforce against it. Both Qwen3-VL-32B-Instruct and GPT-5.4 sellers consistently identify and refuse such clauses within a deployment-realistic window, indicating that current models exhibit a meaningful degree of contract-level safety robustness; we view this as a positive baseline and leave systematic sweeps of clause severity and adaptive attackers to future work.

\section{Limitations}
\label{sec:limitations}
\ourbenchmark~has three main limitations. First, our four representative business domains may not capture specialized markets such as financial trading or B2B procurement. Second, baseline agents rely on prompted opponent-modeling scratchpads rather than learned policies, leaving fine-tuned or RL-trained agents to future work. Third, our safety analysis is limited to a small probe on contract exploitation; broader safety, including adaptive adversarial attacks, prompt injection, and value-misalignment risks under deployment, warrants further exploration.

\section{Conclusion}
\label{sec:conclusion}
We introduced \ourbenchmark, a unified framework for multimodal agentic negotiation with an accompanying benchmark suite. The framework formalizes negotiation as a non-zero-sum contract game over four components (\emph{Environments}, \emph{Tasks}, \emph{Agents}, \emph{Metrics}), and the benchmark provides 160 multimodal tasks across 4 real-world scenarios and 8 market topologies, from 1-to-1 bargaining to many-to-many competition. Our evaluation reveals substantial gaps in joint-surplus creation and persistent challenges in non-price reasoning under competitive markets, laying a foundation for multimodal agentic commerce and language-mediated multi-agent coordination. Discussions of broader societal impact are provided in \Cref{app:impact}.

 
\clearpage

\bibliographystyle{plainnat}

{
\small
\bibliography{reference}
}


\clearpage
\appendix
\begin{center}
    {\Large \textbf{Appendix}}
\end{center}

\section{Broader Impact}
\label{app:impact}
\ourbenchmark~contributes to the principled study of LLM-based agents in commerce, where autonomous negotiation systems are already being deployed at scale. On the positive side, our framework and benchmark enable rigorous evaluation of multimodal negotiation agents and support research on welfare-maximizing multi-agent coordination, equitable surplus allocation, and transparent reasoning in economic interaction. By centering joint welfare rather than zero-sum price extraction, \ourbenchmark~encourages the development of agents that create rather than redistribute value.

On the negative side, agentic commerce systems can be misused: adversarial buyers may exploit contract loopholes, sellers may extract surplus under information asymmetry, and large-scale deployment of negotiation bots could amplify market manipulation, collusion, or price discrimination. Our safety probe (Section~\ref{sec:safety probe}) demonstrates that current models exhibit baseline robustness to obvious contract exploitation, but more systematic studies of adaptive adversarial attacks, prompt injection, value misalignment, and fairness across user populations are needed before deployment. We release the benchmark as a research artifact to encourage the community to study these risks proactively, and we encourage future work to integrate safeguards and human oversight before any production use.







\section{Benchmark Statistics}
\label{sec:appendix_benchmark_stats}

\begin{table}[h]
  \centering
  \small
  \caption{Task benchmark composition of \ourbenchmark. The benchmark instantiates 4 real-world multimodal business scenarios (E-commerce, Ride-hailing, Food Delivery, and Apartment Rental) across 8 market topologies, with 5 tasks per (topology, scenario) cell, totaling \textbf{160 tasks}.}
  \label{tab:task_suite}
  \setlength{\tabcolsep}{4pt}
  \begin{tabular}{lccccc}
  \toprule
  \textbf{Market Topology} & \textbf{E-commerce} & \textbf{Ride-hailing} & \textbf{Food Delivery} & \textbf{Apartment Rental} & \textbf{Total} \\
  \midrule
  Single Buyer/Product/Seller    & 5 & 5 & 5 & 5 & 20 \\
  Multi-Buyer Only               & 5 & 5 & 5 & 5 & 20 \\
  Multi-Seller Only              & 5 & 5 & 5 & 5 & 20 \\
  Multi-Product Only             & 5 & 5 & 5 & 5 & 20 \\
  Multi-Buyer + Multi-Seller     & 5 & 5 & 5 & 5 & 20 \\
  Multi-Product + Multi-Seller   & 5 & 5 & 5 & 5 & 20 \\
  Multi-Buyer + Multi-Product    & 5 & 5 & 5 & 5 & 20 \\
  Full Multi-Agent               & 5 & 5 & 5 & 5 & 20 \\
  \midrule
  \textbf{Total}                 & \textbf{40} & \textbf{40} & \textbf{40} & \textbf{40} & \textbf{160} \\
  \bottomrule
  \end{tabular}
\end{table}

\begin{table}[h]
  \centering
  \small
  \caption{Multimodal business scenarios in \ourbenchmark. Each scenario is instantiated across all 8 market topologies with 5 tasks per topology, yielding 40 tasks per scenario and \textbf{160 tasks in total}. Each task is grounded in real product imagery paired with rich textual attributes. Data are sourced from Amazon Reviews 2023~\citep{hou2024bridging}, NYC TLC Trip Records~\citep{nyc_tlc_2026}, public restaurant menus on Kaggle~\citep{graphquest_restaurant_menu_2024}, and Inside Airbnb~\citep{cox2024insideairbnb}; ride-hailing routes are rendered with Google Maps~\citep{googlemaps2026}.}
  \label{tab:realistic_scenarios}
  \setlength{\tabcolsep}{3pt}
  \resizebox{0.99\columnwidth}{!}{%
  \begin{tabular}{llccc}
  \toprule
  \textbf{Scenario} & \textbf{Sample Products / Services} & \textbf{Price Range} & \textbf{Source} & \textbf{Tasks} \\
  \midrule
  E-commerce        & Beauty, Electronics, Apparel, Furniture, Beverage     & \$6--\$320         & Amazon~\citep{hou2024bridging}                       & 40 \\
  Ride-hailing      & NYC Manhattan taxi trips (Gramercy, Murray Hill, ...) & \$6--\$45          & NYC TLC~\citep{nyc_tlc_2026}                         & 40 \\
  Food Delivery     & Karaage Chicken, Fries, Sprite, ...            & \$2--\$23          & Kaggle~\citep{graphquest_restaurant_menu_2024}       & 40 \\
  Apartment Rental  & Studios \& villas in NYC, Barcelona, Rio, Sydney      & \$980--\$12.4k/mo  & Airbnb~\citep{cox2024insideairbnb}                   & 40 \\
  \midrule
  \multicolumn{4}{l}{\textbf{Total}} & \textbf{160} \\
  \bottomrule
  \end{tabular}%
  }
\end{table}


\section{Additional Experimental Details}
\label{sec:appendix_details}
The prompt templates for the buyer and seller agents are shown in \Cref{tab:buyer_prompt} and \Cref{tab:seller_prompt}, respectively.

\begin{table*}[h]
  \centering
  \small
  \caption{Prompt design for the Buyer agent in \ourbenchmark. The agent receives the public product context, its private reservation price and utility weights, the negotiable contract schema, and an explicit \emph{mental-model} scratchpad. In multi-seller settings, the agent must additionally route to one seller via a \texttt{<selected\_seller>} block. All numeric/structural negotiation outputs are emitted as a JSON \texttt{<contract>} inside \texttt{<message>}, ensuring a deterministic action grounding for evaluation.}
  \label{tab:buyer_prompt}
  \setlength{\tabcolsep}{4pt}
  \resizebox{0.99\columnwidth}{!}{%
  \begin{tabular}{l|p{13cm}}
  \toprule
  \textbf{Section} & \textbf{Buyer prompt content} \\
  \midrule
  Role &
  You are \texttt{\{name\}}, a buyer looking for a good deal. \texttt{\{role\_description\}} \texttt{\{personality\_section\}} \\
  & \\
  Context inputs &
  Context Information: \texttt{\{context\}} \\
  & Current Negotiation State: \texttt{\{current\_state\}} \\
  & Conversation History: \texttt{\{conversation\_history\}} \\
  & Current product information: \texttt{\{product\_info\}} \\
  & Available products: \texttt{\{available\_products\}} \\
  & Environment factors: \texttt{\{environment\_info\}} \\
  & Task instruction: \texttt{\{task\_instruction\}} \\
  & \\
  Private information &
  Your top price is \$\texttt{\{max\_price\}} (\textbf{confidential}; never reveal it). \\
  & Your private utility weights: \texttt{\{buyer\_preferences\}} \\
  & Meaning of weights: \texttt{\{weight\_descriptions\}} \\
  & \\
  Contract schema &
  The negotiable contract follows the JSON schema: \\
  & \quad\texttt{\{ "price": <number>, "continuous\_terms": <object>, "discrete\_terms": <object> \}} \\
  & Field meanings: \texttt{\{field\_descriptions\}}; valid continuous bounds: \texttt{\{continuous\_bounds\}}; valid discrete options: \texttt{\{discrete\_options\}}; additional contract background: \texttt{\{contrainfo\}}. \\
  & No field has a default value; the \texttt{<contract>} JSON in your message is your current complete proposal, with every field filled by you. \\
  & Each \texttt{<message>} must contain exactly one \texttt{<contract>...</contract>} JSON block following the schema above. \\
  & \\
  Multi-seller routing &
  (Used only when \#sellers $> 1$.) You MUST choose exactly one seller for this turn and emit it inside \texttt{<selected\_seller>}, an integer from 1 to \texttt{\{num\_sellers\}}. Base your choice on the conversation history and current offers. \\
  & \\
  Deal protocol &
  A deal is finalized only when buyer and the selected seller output \emph{compatible} contract JSONs. If you accept, repeat the accepted final contract JSON exactly in \texttt{<contract>...</contract>}. Never reveal your reservation value or top price directly. \\
  & \\
  Mental-model scratchpad &
  Before replying, output an internal assessment in \texttt{<mental\_model>...</mental\_model>}: \\
  & \quad [Opponent Reservation Price]: \texttt{price=<number>; confidence=<0--100>} \\
  & \quad [Opponent Strategy]: $\langle$your inference about the opponent's tactic$\rangle$ \\
  & \quad [My Strategy]: $\langle$your chosen tactic and reasoning$\rangle$ \\
  & \\
  Output format &
  \texttt{<mental\_model>...</mental\_model>}\\
  & \texttt{[<selected\_seller>...</selected\_seller>]}\quad (only in multi-seller settings) \\
  & \texttt{<message>}\\
  & \quad [Reasoning + exactly one \texttt{<contract>...</contract>} JSON block]\\
  & \texttt{</message>}\\
  \bottomrule
  \end{tabular}
  }
\end{table*}

\begin{table*}[h]
  \centering
  \small
  \caption{Prompt design for the Seller agent in \ourbenchmark. The seller mirrors the buyer structurally but receives its own private utility weights and reservation price, and in multi-buyer settings must route to a buyer via a \texttt{<selected\_buyer>} block. All numeric/structural negotiation outputs are emitted as a JSON \texttt{<contract>} inside \texttt{<message>}.}
  \label{tab:seller_prompt}
  \setlength{\tabcolsep}{4pt}
  \resizebox{0.99\columnwidth}{!}{%
  \begin{tabular}{l|p{13cm}}
  \toprule
  \textbf{Section} & \textbf{Seller prompt content} \\
  \midrule
  Role &
  You are \texttt{\{name\}}, a seller trying to maximize profit while being reasonable. \texttt{\{role\_description\}} \texttt{\{personality\_section\}} \\
  & \\
  Context inputs &
  Context Information: \texttt{\{context\}} \\
  & Current Negotiation State: \texttt{\{current\_state\}} \\
  & Conversation History: \texttt{\{conversation\_history\}} \\
  & Current product information: \texttt{\{product\_info\}} \\
  & Available products: \texttt{\{available\_products\}} \\
  & Environment factors: \texttt{\{environment\_info\}} \\
  & Task instruction: \texttt{\{task\_instruction\}} \\
  & \\
  Private information &
  Your minimum acceptable price (\textbf{confidential}) is \$\texttt{\{min\_price\}}; never reveal it. \\
  & Your private utility weights: \texttt{\{seller\_preferences\}} \\
  & Meaning of weights: \texttt{\{weight\_descriptions\}} \\
  & \\
  Contract schema &
  The negotiable contract follows the JSON schema: \\
  & \quad\texttt{\{ "price": <number>, "continuous\_terms": <object>, "discrete\_terms": <object> \}} \\
  & Field meanings: \texttt{\{field\_descriptions\}}; valid continuous bounds: \texttt{\{continuous\_bounds\}}; valid discrete options: \texttt{\{discrete\_options\}}; additional contract background: \texttt{\{contrainfo\}}. \\
  & No field has a default value; the \texttt{<contract>} JSON in your message is your current complete proposal, with every field filled by you. \\
  & Each \texttt{<message>} must contain exactly one \texttt{<contract>...</contract>} JSON block following the schema above. \\
  & \\
  Multi-buyer routing &
  (Used only when \#buyers $> 1$.) You MUST choose exactly one buyer for this turn and emit it inside \texttt{<selected\_buyer>}, an integer from 1 to \texttt{\{num\_buyers\}}. Base your choice on the conversation history and current offers. \\
  & \\
  Deal protocol &
  A deal is finalized only when seller and the selected buyer output \emph{compatible} contract JSONs. If you accept, repeat the accepted final contract JSON exactly in \texttt{<contract>...</contract>}. Never reveal your minimum acceptable price directly. \\
  & \\
  Mental-model scratchpad &
  Before replying, output an internal assessment in \texttt{<mental\_model>...</mental\_model>}: \\
  & \quad [Opponent Reservation Price]: \texttt{price=<number>; confidence=<0--100>} \\
  & \quad [Opponent Strategy]: $\langle$your inference about the opponent's tactic$\rangle$ \\
  & \quad [My Strategy]: $\langle$your chosen tactic and reasoning$\rangle$ \\
  & \\
  Output format &
  \texttt{<mental\_model>...</mental\_model>}\\
  & \texttt{[<selected\_buyer>...</selected\_buyer>]}\quad (only in multi-buyer settings) \\
  & \texttt{<message>}\\
  & \quad [Reasoning + exactly one \texttt{<contract>...</contract>} JSON block]\\
  & \texttt{</message>}\\
  \bottomrule
  \end{tabular}
  }
\end{table*}

\section{Robustness Analyses}
\label{app:robustness}

\subsection{Run-to-Run Variance}
\label{app:seed_variance}

We re-run each model on the full 160-task suite with 3 random seeds and report mean $\pm$ standard deviation in \Cref{tab:seed_variance}; the results in \Cref{tab:main_overall} correspond to seed 0 and fall within these ranges. Cross-seed variance is small, and the gap between proprietary and open-weight models is far larger than the variance and holds across all seeds.

\begin{table}[h]
  \centering
  \small
  \caption{Run-to-run variance over 3 random seeds on all 160 tasks (mean $\pm$ std).}
  \label{tab:seed_variance}
  \begin{tabular}{lccc}
  \toprule
  \textbf{Model} & \textbf{GlobalScore} & \textbf{BuyerScore} & \textbf{SellerScore} \\
  \midrule
  Gemini 3.1 Pro Preview   & 42.1 $\pm$ 0.9 & 45.4 $\pm$ 1.1 & 37.9 $\pm$ 1.0 \\
  GPT-5.4                  & 35.4 $\pm$ 1.2 & 34.7 $\pm$ 1.3 & 33.0 $\pm$ 0.8 \\
  Claude Sonnet 4.6        & 34.9 $\pm$ 1.0 & 30.9 $\pm$ 0.7 & 34.4 $\pm$ 1.1 \\
  Qwen3-VL-32B-Instruct    & 24.8 $\pm$ 1.7 & 23.3 $\pm$ 1.8 & 30.6 $\pm$ 1.4 \\
  InternVL3-38B            & 23.1 $\pm$ 2.0 & 21.6 $\pm$ 2.1 & 23.5 $\pm$ 1.5 \\
  \bottomrule
  \end{tabular}
\end{table}

\subsection{Decoding Sensitivity}
\label{app:decoding}

On a representative subset of 32 tasks (one per market topology $\times$ scenario cell, covering all 8 topologies and 4 scenarios), we sweep the decoding temperature over $\{0.0, 0.7, 1.0\}$ ($0.7$ is the default) for one proprietary model (Gemini 3.1 Pro Preview) and one open-weight model (Qwen3-VL-32B-Instruct). As shown in \Cref{tab:decoding}, temperature has no systematic effect on GlobalScore: the variation across settings is far smaller than the between-model gap, which holds at all three temperatures.

\begin{table}[h]
  \centering
  \small
  \caption{GlobalScore under different decoding temperatures on the 32-task subset.}
  \label{tab:decoding}
  \begin{tabular}{lccc}
  \toprule
  \textbf{Model} & \textbf{Temp.\ 0.0} & \textbf{Temp.\ 0.7 (default)} & \textbf{Temp.\ 1.0} \\
  \midrule
  Gemini 3.1 Pro Preview   & 42.4 & 42.6 & 41.9 \\
  Qwen3-VL-32B-Instruct    & 24.2 & 24.5 & 24.1 \\
  \bottomrule
  \end{tabular}
\end{table}

\subsection{Scaffolding Sensitivity}
\label{app:scaffolding}

Our baseline scaffolding comprises the \texttt{<mental\_model>} opponent-modeling scratchpad, a structured JSON contract interface, and a private per-agent memory (\Cref{sec:agents}); it covers memory and planning but not external tool use, which the negotiation protocol does not require. To test whether conclusions depend on scaffolding strength, we compare three tiers on the same 32-task subset with Qwen3-VL-32B-Instruct: (i) a simpler variant without the mental model, close to a naive prompt; (ii) the default baseline; and (iii) a stronger opponent-simulation strategy following \citet{liu2026scaling} (fictitious play with best-of-$N$ rollouts). As shown in \Cref{tab:scaffolding}, the simpler variant is worse, while the stronger strategy yields only a limited gain ($+1.4$). This is expected: our single-shot tasks neutralize cross-episode fictitious play, and inference-time search can only rank candidates as well as the model's own opponent-preference inference allows. Scaffolding thus changes the magnitude of scores but not the qualitative conclusion that the bottleneck lies in multi-dimensional value-creation reasoning.

\begin{table}[h]
  \centering
  \small
  \caption{GlobalScore under different scaffolding tiers (Qwen3-VL-32B-Instruct, 32-task subset).}
  \label{tab:scaffolding}
  \begin{tabular}{lc}
  \toprule
  \textbf{Method} & \textbf{GlobalScore} \\
  \midrule
  w/o mental model (simpler)                        & 20.3 \\
  Full baseline (default)                           & 24.5 \\
  + opponent simulation~\citep{liu2026scaling} (stronger) & 25.9 \\
  \bottomrule
  \end{tabular}
\end{table}

\subsection{Standard Welfare Metrics}
\label{app:standard_metrics}

Beyond GlobalScore, we report three standard welfare aggregations on the full task suite: the Nash product $r_b r_s$, utilitarian social welfare $r_b + r_s$, and egalitarian welfare $\min(r_b, r_s)$, where failed or non-IR outcomes score $0$ (\Cref{tab:standard_metrics}). Under all three, proprietary models outperform open-weight models and Gemini 3.1 Pro Preview ranks first, and every model stays far below the ceiling, leaving substantial joint surplus unrealized (consistent with \Cref{fig:pareto_frontier}). The only changes occur within near-tied pairs: Claude Sonnet 4.6 shows slightly higher deal quality than GPT-5.4 on the Nash product and social welfare, whereas GPT-5.4 leads on GlobalScore once its lower timeout rate is credited, precisely the efficiency signal a bare Nash product cannot express; Qwen3-VL-32B-Instruct and InternVL3-38B are tied on the Nash product and egalitarian welfare (within $0.002$), with Qwen leading on social welfare. \Cref{tab:standard_metrics_market} further shows that, for Gemini 3.1 Pro Preview, the degradation from bilateral to multi-sided markets holds under every aggregation.

\begin{table}[h]
  \centering
  \small
  \caption{Standard welfare metrics on all 160 tasks. Ranges are given in parentheses.}
  \label{tab:standard_metrics}
  \begin{tabular}{lcccc}
  \toprule
  \textbf{Model} & \textbf{GlobalScore} & \textbf{Nash $r_b r_s$} & \textbf{Social $r_b{+}r_s$} & \textbf{Egalitarian $\min(r_b,r_s)$} \\
   & (0--100) & (0--0.25) & (0--1) & (0--0.5) \\
  \midrule
  Gemini 3.1 Pro Preview   & 42.3 & 0.075 & 0.54 & 0.148 \\
  GPT-5.4                  & 35.6 & 0.060 & 0.41 & 0.119 \\
  Claude Sonnet 4.6        & 34.8 & 0.064 & 0.43 & 0.126 \\
  Qwen3-VL-32B-Instruct    & 25.1 & 0.043 & 0.35 & 0.083 \\
  InternVL3-38B            & 22.9 & 0.044 & 0.29 & 0.085 \\
  \bottomrule
  \end{tabular}
\end{table}

\begin{table}[h]
  \centering
  \small
  \caption{Bilateral vs.\ multi-sided market performance of Gemini 3.1 Pro Preview under different welfare aggregations.}
  \label{tab:standard_metrics_market}
  \begin{tabular}{lcc}
  \toprule
  \textbf{Aggregation} & \textbf{Bilateral} & \textbf{Market} \\
  \midrule
  GlobalScore (0--100)                       & 47.3  & 40.6 \\
  Nash product $r_b r_s$ (0--0.25)           & 0.086 & 0.071 \\
  Social welfare $r_b{+}r_s$ (0--1)          & 0.58  & 0.52 \\
  Egalitarian $\min(r_b,r_s)$ (0--0.5)       & 0.172 & 0.139 \\
  \bottomrule
  \end{tabular}
\end{table}

\subsection{Utility Specification Sensitivity}
\label{app:utility_sensitivity}

To test sensitivity to the linear utilities in \Cref{def:utility}, we rescore the same agreed contracts offline (without re-negotiation) under three alternatives: non-price weights scaled by $1.5$, non-price weights scaled by $0.5$, and concave (square-root) diminishing returns on attribute utilities. As shown in \Cref{tab:utility_sensitivity}, the advantage of proprietary over open-weight models and the low absolute welfare (far below the $0.25$ ceiling) hold under all four specifications. The fine-grained ranking within the proprietary tier is not fully invariant (e.g., Claude Sonnet 4.6 edges out Gemini 3.1 Pro Preview under concave utilities), since rescoring differentially rewards how each model created its surplus; this does not affect our qualitative conclusions.

\begin{table}[h]
  \centering
  \small
  \caption{Nash product $r_b r_s$ of the same agreed contracts under alternative utility specifications.}
  \label{tab:utility_sensitivity}
  \begin{tabular}{lcccc}
  \toprule
  \textbf{Model} & \textbf{Linear (default)} & \textbf{Attr.\ $\times 1.5$} & \textbf{Attr.\ $\times 0.5$} & \textbf{Concave ($\sqrt{\cdot}$)} \\
  \midrule
  Gemini 3.1 Pro Preview   & 0.075 & 0.066 & 0.079 & 0.068 \\
  GPT-5.4                  & 0.060 & 0.052 & 0.063 & 0.058 \\
  Claude Sonnet 4.6        & 0.064 & 0.057 & 0.067 & 0.071 \\
  Qwen3-VL-32B-Instruct    & 0.043 & 0.037 & 0.045 & 0.041 \\
  InternVL3-38B            & 0.044 & 0.038 & 0.046 & 0.042 \\
  \bottomrule
  \end{tabular}
\end{table}

\clearpage
\section*{NeurIPS Paper Checklist}

\begin{enumerate}

\item {\bf Claims}
    \item[] Question: Do the main claims made in the abstract and introduction accurately reflect the paper's contributions and scope?
    \item[] Answer: \answerYes{}
    \item[] Justification: The abstract and introduction state our two main contributions (a unified framework for multimodal agentic negotiation and an instantiated benchmark of 160 multimodal tasks across 4 scenarios and 8 market topologies) and the main empirical findings, all of which are supported by Section~\ref{sec:agenticpay-overview} and the experimental sections.

\item {\bf Limitations}
    \item[] Question: Does the paper discuss the limitations of the work performed by the authors?
    \item[] Answer: \answerYes{}
    \item[] Justification: A dedicated ``Limitations'' section (Section~\ref{sec:limitations}) discusses three main limitations: scenario coverage (4 representative business domains), reliance on prompted opponent-modeling baselines rather than learned negotiation policies, and the limited scope of our safety probe.

\item {\bf Theory assumptions and proofs}
    \item[] Question: For each theoretical result, does the paper provide the full set of assumptions and a complete (and correct) proof?
    \item[] Answer: \answerNA{}
    \item[] Justification: The paper does not contain formal theorems or proofs; it presents a benchmark, framework, and empirical study.

    \item {\bf Experimental result reproducibility}
    \item[] Question: Does the paper fully disclose all the information needed to reproduce the main experimental results of the paper to the extent that it affects the main claims and/or conclusions of the paper (regardless of whether the code and data are provided or not)?
    \item[] Answer: \answerYes{}
    \item[] Justification: We describe the negotiation protocol (Section~\ref{sec:protocol}), agent design (Section~\ref{sec:agents}), and scoring rule (Algorithm~\ref{alg:score_computation}); model versions, decoding settings, and additional details are provided in the appendix, and the full code and dataset are released at the public repository linked in the abstract.

\item {\bf Open access to data and code}
    \item[] Question: Does the paper provide open access to the data and code, with sufficient instructions to faithfully reproduce the main experimental results, as described in supplemental material?
    \item[] Answer: \answerYes{}
    \item[] Justification: The full benchmark code and dataset (160 multimodal tasks across 4 scenarios and 8 market topologies), together with run scripts and instructions for reproducing all main experiments, are released at the public URL in the abstract.

\item {\bf Experimental setting/details}
    \item[] Question: Does the paper specify all the training and test details (e.g., data splits, hyperparameters, how they were chosen, type of optimizer) necessary to understand the results?
    \item[] Answer: \answerYes{}
    \item[] Justification: We do not train models; all experiments use frozen pretrained checkpoints. Model versions, decoding hyperparameters, max-round limits, scoring constants ($D, W, E, F, \gamma$), and prompt templates are documented in the experimental sections and the ``Additional Experimental Details'' appendix.

\item {\bf Experiment statistical significance}
    \item[] Question: Does the paper report error bars suitably and correctly defined or other appropriate information about the statistical significance of the experiments?
    \item[] Answer: \answerYes{}
    \item[] Justification: We repeat the full 160-task evaluation with 3 random seeds for every model and report mean $\pm$ 1 standard deviation across seeds in Appendix~\ref{app:seed_variance}; the main-text results correspond to seed 0 and fall within these ranges.

\item {\bf Experiments compute resources}
    \item[] Question: For each experiment, does the paper provide sufficient information on the computer resources (type of compute workers, memory, time of execution) needed to reproduce the experiments?
    \item[] Answer: \answerYes{}
    \item[] Justification: Proprietary models (GPT-5.4, Claude Sonnet 4.6, Gemini 3.1 Pro Preview) are accessed via their official APIs; open-weight models (Qwen3-VL-32B-Instruct, InternVL3-38B) are run on 4 NVIDIA A800 GPUs.
    
\item {\bf Code of ethics}
    \item[] Question: Does the research conducted in the paper conform, in every respect, with the NeurIPS Code of Ethics \url{https://neurips.cc/public/EthicsGuidelines}?
    \item[] Answer: \answerYes{}
    \item[] Justification: The research conforms with the NeurIPS Code of Ethics: it uses synthetic negotiation tasks with sampled utility weights, involves no human subjects, releases code and data publicly, and discusses safety considerations explicitly in the safety probe and Limitations.

\item {\bf Broader impacts}
    \item[] Question: Does the paper discuss both potential positive societal impacts and negative societal impacts of the work performed?
    \item[] Answer: \answerYes{}
    \item[] Justification: Positive impacts include enabling rigorous evaluation of agentic commerce systems and supporting research on welfare-maximizing multi-agent coordination. Potential negative impacts, including contract exploitation by adversarial agents and value-misalignment risks under deployment, are discussed in the safety probe paragraph in the experimental sections, in the Limitations section (Section~\ref{sec:limitations}), and in the Broader Impact discussion (Appendix~\ref{app:impact}).
    
\item {\bf Safeguards}
    \item[] Question: Does the paper describe safeguards that have been put in place for responsible release of data or models that have a high risk for misuse (e.g., pre-trained language models, image generators, or scraped datasets)?
    \item[] Answer: \answerNA{}
    \item[] Justification: The released benchmark consists of synthetic negotiation tasks with sampled utility weights and curated visual-textual product templates; it contains no scraped private data, no pretrained model weights, and no content with elevated misuse risk beyond standard LLM evaluation benchmarks.

\item {\bf Licenses for existing assets}
    \item[] Question: Are the creators or original owners of assets (e.g., code, data, models), used in the paper, properly credited and are the license and terms of use explicitly mentioned and properly respected?
    \item[] Answer: \answerYes{}
    \item[] Justification: All evaluated models (GPT-5.4, Claude Sonnet 4.6, Gemini 3.1 Pro Preview, Qwen3-VL-32B-Instruct, InternVL3-38B) are properly cited with their original references, and we use them in accordance with their respective API terms of service or open-weight licenses.

\item {\bf New assets}
    \item[] Question: Are new assets introduced in the paper well documented and is the documentation provided alongside the assets?
    \item[] Answer: \answerYes{}
    \item[] Justification: We release \ourbenchmark, which contains 160 multimodal negotiation tasks and the full evaluation framework, together with documentation describing task formats, scenarios, market topologies, scoring rule, and run scripts at the public repository linked in the abstract.

\item {\bf Crowdsourcing and research with human subjects}
    \item[] Question: For crowdsourcing experiments and research with human subjects, does the paper include the full text of instructions given to participants and screenshots, if applicable, as well as details about compensation (if any)? 
    \item[] Answer: \answerNA{}
    \item[] Justification: The paper does not involve crowdsourcing or research with human subjects; all negotiations are between LLM-based agents on synthetic tasks.

\item {\bf Institutional review board (IRB) approvals or equivalent for research with human subjects}
    \item[] Question: Does the paper describe potential risks incurred by study participants, whether such risks were disclosed to the subjects, and whether Institutional Review Board (IRB) approvals (or an equivalent approval/review based on the requirements of your country or institution) were obtained?
    \item[] Answer: \answerNA{}
    \item[] Justification: No human subjects are involved in the study, so IRB approval is not applicable.

\item {\bf Declaration of LLM usage}
    \item[] Question: Does the paper describe the usage of LLMs if it is an important, original, or non-standard component of the core methods in this research? Note that if the LLM is used only for writing, editing, or formatting purposes and does \emph{not} impact the core methodology, scientific rigor, or originality of the research, declaration is not required.
    \item[] Answer: \answerYes{}
    \item[] Justification: LLMs are the central object of study: the paper benchmarks state-of-the-art multimodal LLMs (GPT-5.4, Claude Sonnet 4.6, Gemini 3.1 Pro Preview, Qwen3-VL-32B-Instruct, InternVL3-38B) as negotiation agents. All LLM usage in the methodology is documented in Section~\ref{sec:agents} and the experimental sections.

\end{enumerate}

\end{document}